\documentclass[10pt,twocolumn,letterpaper]{article}

\usepackage[pagenumbers]{arxiv} % To force page numbers, e.g. for an arXiv version

\definecolor{arxivblue}{rgb}{0.21,0.49,0.74}
\usepackage[pagebackref,breaklinks,colorlinks,allcolors=arxivblue]{hyperref}

\usepackage{colortbl}
\usepackage{pifont}
\newcommand{\bftab}{\fontseries{b}\selectfont}
\usepackage{multirow}

\title{Token Transforming: A Unified and Training-Free Token Compression Framework for Vision Transformer Acceleration}

\author{Fanhu Zeng$^{1}$, 
    Deli Yu$^{1}$, 
    Zhenglun Kong$^{2}$,
    Hao Tang$^{3}$ \\
    $^{1}$ Institute of Automation, Chinese Academy of Sciences \\
    $^{2}$ Harvard University \quad $^{3}$School of Computer Science, Peking University\\
}

\begin{document}
\maketitle
\begin{abstract}
Vision transformers have been widely explored in various vision tasks. Due to heavy computational cost, much interest has aroused for compressing vision transformer dynamically in the aspect of tokens. Current methods mainly pay attention to token pruning or merging to reduce token numbers, in which tokens are compressed exclusively, causing \textbf{great information loss} and therefore \textbf{post-training is inevitably required} to recover the performance. In this paper, we rethink token reduction and \textbf{unify the process as an explicit form of token matrix transformation}, in which all existing methods are constructing special forms of matrices within the framework. Furthermore, we propose a many-to-many Token Transforming framework that serves as a generalization of all existing methods and reserves the most information, even enabling \textbf{training-free} acceleration. We conduct extensive experiments to validate our framework. Specifically, we reduce \textbf{40\% FLOPs} and accelerate DeiT-S by \textbf{$\times$1.5} with marginal 0.1\% accuracy drop. Furthermore, we extend the method to dense prediction tasks including segmentation, object detection, depth estimation, and language model generation. Results demonstrate that the proposed method consistently achieves substantial improvements, offering a better computation-performance trade-off, impressive budget reduction and inference acceleration.
\end{abstract}

\section{Introduction}
\label{sec:intro}
Research on Vision Transformers~\cite{dosovitskiy2020image} has made breakthrough in various downstream vision tasks including image classification~\cite{touvron2021training, yuan2021tokens}, object detection~\cite{li2022exploring, zhang2022dino}, semantic segmentation~\cite{strudel2021segmenter} and so on \cite{{ryoo2021tokenlearner,lee2022mpvit}}. However, quadratic computation in proportion to the number of tokens significantly prevents wide application~\cite{ranftl2021vision}. To this end, model compression~\cite{gao2018dynamic, hinton2015distilling, sui2021chip, wang2022qsfm, wang2021not} is proposed to reduce redundant computation within the model, and pruning~\cite{molchanov2019importance, yang2023global} has become common practice for compression to improve real-time efficiency. Recent research focuses on pruning tokens in a dynamic way~\cite{liu2022adaptive} as it is consistent with common sense of human cognition that both important attentive regions and neglected uninformative areas dynamically vary with given images to obtain better accuracy-efficiency trade-off~\cite{han2021dynamic}. Pioneering works~\cite{rao2021dynamicvit, fayyaz2022adaptive, xu2022evo, liang2022not} prune tokens with low importance score directly, which is calculated by trainable prediction module or statistics of attention maps. Considering information loss during pruning, others~\cite{bolya2022token, wei2023joint, kong2022spvit,  long2023beyond, li2024vidtome} adopt token merging, which merges them into informative tokens or cluster them into fewer representative groups and achieve impressive promotion.

Despite great progress, critical problems still remain unsolved. \textbf{(1) Tokens to be merged are exclusive.} In other words, if a token is assigned to a certain group, it cannot be assigned to other groups again, which greatly restricts the flexibility of information expression, as crucial tokens may be attached to more than one token at certain moments. \textbf{(2) Due to severe accuracy drop, post-training is still required} to recover performance~\cite{liang2022not, rao2021dynamicvit, long2023beyond} , which is resource-intensive and constraints their practical application. Although some~\cite{bolya2022token,wu2023ppt,liang2022expediting} claim training-free compression, the acceleration is limited. 

\begin{figure}[tb]
\centering
\includegraphics[width=\linewidth]{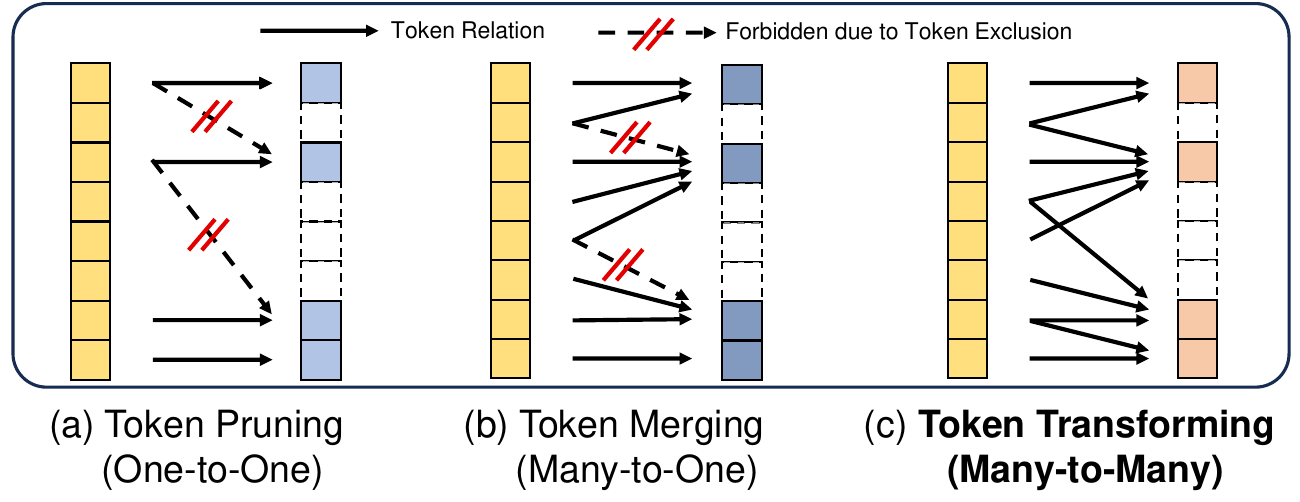}
\vspace{-15pt}
\caption{\footnotesize Comparison of different token reduction between original~(left column) and remaining~(right column) tokens. (a) Pruning methods collect tokens from corresponding one token~(one-to-one). (b) Merging methods exclusively merge multiple original tokens into one remaining token~(many-to-one). (c) Token Transforming integrates original tokens into remaining ones by matrix transformation~(many-to-many).}
\label{fig:main-structure}
\vspace{-15pt}
\end{figure}

\begin{figure}[tbp]
  \centering
  \includegraphics[width=\linewidth]{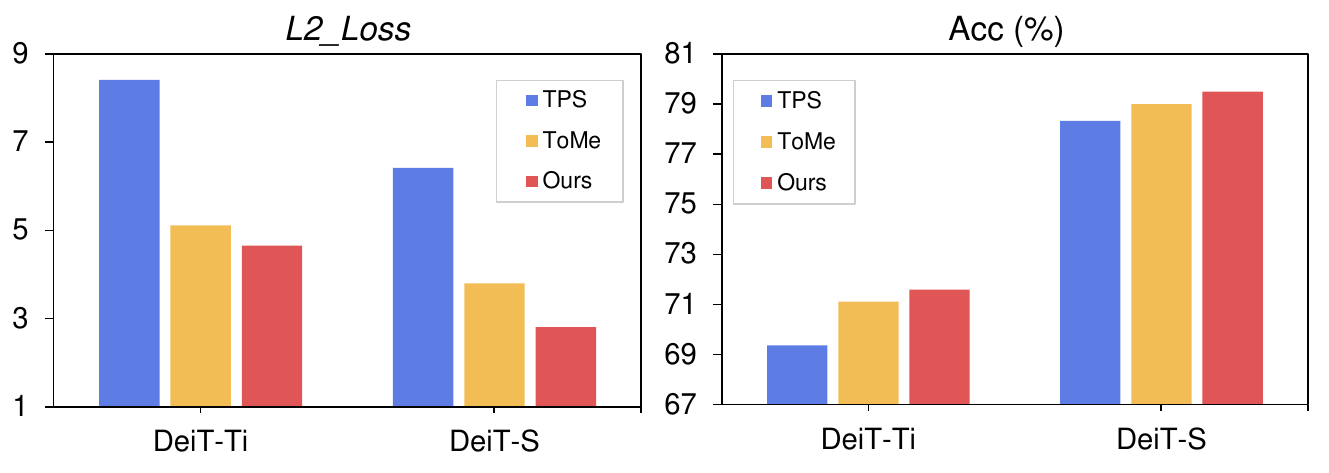}
  \vspace{-20pt}
  \caption{Relationship between the error of class token and the accuracy in image classification. The trend is that the less token information loss, the higher the accuracy. Our approach reserves information to the most and achieves the best accuracy. }
  \vspace{-5pt}
\label{fig:bias_descrption}
\end{figure}

The above analysis naturally motivates us to an open question: \textit{can we unify the two types of token reduction to achieve high-performance model compression?} We start from general expression of token pruning and merging. As is shown in~\cref{fig:main-structure}\textcolor{arxivblue}{a} and~\cref{fig:main-structure}\textcolor{arxivblue}{b}, we discover that token reduction has a similar mapping relationship, which can be regarded as token matrix transformation. From this perspective, we find that the two type of methods can be seen as special, token-exclusive cases of transformation matrix with diagonal-wise and block-wise form of transformation matrix, respectively. The reason existing methods cannot achieve satisfactory compression may result from special, exclusive and sub-optimal transformation matrix, which can be validated by class token error and accuracy comparison shown in~\cref{fig:bias_descrption} that finer transformation design can achieve higher accuracy with lower information loss.

Motivated by this, we propose a unified Token Transformation framework to overcome the limitation of existing token-exclusive reduction methods, in which originally tokens are transformed in a more flexible \textbf{many-to-many} manner and can be integrated into more than one crucial remaining token shown in~\cref{fig:main-structure}\textcolor{arxivblue}{c}. Built upon this, we further determine a transformation matrix that can reserve token information to the most in a \textbf{training-free} manner. Concretely, we first proposed an attentive-based token selection strategy that selects informative tokens. Next, informative tokens are used to calculate similarity between original tokens and the matrix is determined accordingly. Note that the transformation matrix serves as a generalization of previous work, as it degenerates to diagonal or block-wise one if each token is exclusively assigned in the reduction process.

Comprehensive experiments are carried out to demonstrate the effectiveness and transferability of the framework. For example, we achieve $\times$1.5 training-free acceleration with negligible 0.1\% accuracy drop on DeiT-S, even outperforming existing methods which require fine-tuning. 
Furthermore, we compress models in dense prediction tasks and obtain competitive results. For instance, Segmenter-L can be accelerated by 30.4\% with trivial 0.2 mIoU drop. Results on multimodal language model generation also certify its effectiveness in large models. 

Our contribution is summarized as follows:
\begin{itemize}
\item{We unify token reduction process as an explicit form of token matrix transformation, and propose a unified token transformation framework, in which all existing methods are only special forms of transformation matrices.}
\item{We develop an effective algorithm of expressing transformation matrix, which enables more flexible many-to-many transformation and can reserve token information to the most in a training-free manner.}
\item{We conduct various experiments on image classification, dense prediction tasks including segmentation, object detection, depth estimation and large language model generation with competitive results and substantial inference acceleration to validate the superiority and efficiency.}
\end{itemize}

\section{Related Work}
\label{sec:related_work}

\subsection{Efficient Vision Transformers}
Vision Transformer~\cite{dosovitskiy2020image} successfully demonstrates superior results of Transformers~\cite{vaswani2017attention, radford2018improving} in numerous visual tasks. Due to high training consumption on millions of data~\cite{sun2017revisiting}, various research has been carried out to explore efficient vision transformers~\cite{yuan2021tokens, chen2021crossvit, ren2022beyond, liu2021swin}. DeiT~\cite{touvron2021training} achieves competitive performance through an efficient training paradigm under distillation knowledge. LV-ViT~\cite{jiang2021all} generates a dense score map to extract rich local information. However, their goal is to achieve better performance with less data or few additional modules, and treat all inputs as the same, wasting unnecessary training resources when the image is easy to recognize. By contrast, our method works on a orthogonal direction and intends to reduce amount of calculation reusing trained-weights at fingertips.

\subsection{Dynamic Vision Transformers}
Due to high computational cost of vision transformers, many attempts are made to reduce tokens dynamically according to input content rather than designing efficient structures~\cite{zeng2024m2m,kong2025token}. Token compression strategy can mainly be divided into token pruning and token merging.

\noindent \textbf{Token Pruning} discards uninformative tokens directly, which is a straightforward way for adaptive compression. Different importance assessments are carried out to prune unnecessary parts dynamically according to complexity of input images~\cite{fayyaz2022adaptive, yin2022vit, tang2022patch, kong2023peeling}. DynamicViT~\cite{rao2021dynamicvit} designs a lightweight prediction module to effectively prune redundant tokens. ATS~\cite{fayyaz2022adaptive} proposes an adaptive token sampling method to sample tokens dynamically during inference. Nevertheless, simple pruning suffers from severe accuracy drop due to direct loss of information in images.

\noindent \textbf{Token Merging} combines tokens together to reserve more information~\cite{liang2022not, wei2023joint, long2023beyond}. EViT~\cite{liang2022not} reduces tokens by measuring attention with class token to identify attentive token and fuses inattentive ones dynamically. ToMe~\cite{bolya2022token} incorporates a bipartite matching process to combine tokens according to similarity. TPS~\cite{wei2023joint} squeezes tokens into several reserved ones via exclusive matching. However, these methods merge tokens exclusively, restricting flexibility and utilization of information. By contrast, we incorporate more flexible many-to-many transforming for compression.

\section{Method}
\label{sec:method}
In this section, we first give an overall analysis on transforming in~\cref{sec:overview}. Next, we introduce our proposed method in detail in~\cref{sec:token_definition}-\ref{sec:scaling}. Then, we describe extending our method to dense prediction tasks in~\cref{sec:adapt_dense_prediction}.  

\subsection{Overview of Transformation Matrix}
\label{sec:overview}
Existing token pruning and merging methods have similar token-exclusive mapping relation. We rethink token reduction and discover that the process can be regarded as a general matrix transformation. Formally, the equation of token reduction can be formulated as:
\begin{equation}
    \boldsymbol{Y} = \boldsymbol{W}\cdot\boldsymbol{X},
\end{equation}
where $\boldsymbol{Y} \in \mathbb{R}^{M\times d}$, $\boldsymbol{X}\in \mathbb{R}^{N\times d}$~($M$\textless $N$) stand for tokens before and after transforming, and transformation matrix $\boldsymbol{W} \in \mathbb{R}^{M\times N}$ represents mapping relationship in token reduction. 
For example, token pruning simply discards uninformative tokens and retains only a subset of the original tokens, as shown in~\cref{fig:main-structure}\textcolor{arxivblue}{a}. As a result, the transformation matrix takes a diagonal form with elements of either zero or one on the main diagonal, representing a one-to-one mapping. Token merging, on the other hand, combines multiple tokens into a single token, ensuring that merged tokens are not assigned to other tokens any more. Thus, the transformation matrix is block-wise representing many-to-one token relationship, as shown in~\cref{fig:main-structure}\textcolor{arxivblue}{b}. In summary, prior works can be seen as constructing specialized forms of the transformation matrix, with token-exclusion restricting its flexibility and causing severe information loss.

Built upon this, we propose a unified token reduction framework called Token Transforming, which overcomes the token-exclusive limitation of transformation matrix in existing methods. 
Particularly, we design an effective transformation matrix that enables more flexible many-to-many token transformation, where original tokens can be integrated into more than one crucial remaining token, as shown in~\cref{fig:main-structure}\textcolor{arxivblue}{c}. It can thus reserve token information to the most in a training-free manner during token reduction process. 

\begin{figure*}[t]
\centering
\includegraphics[width=2.0\columnwidth]{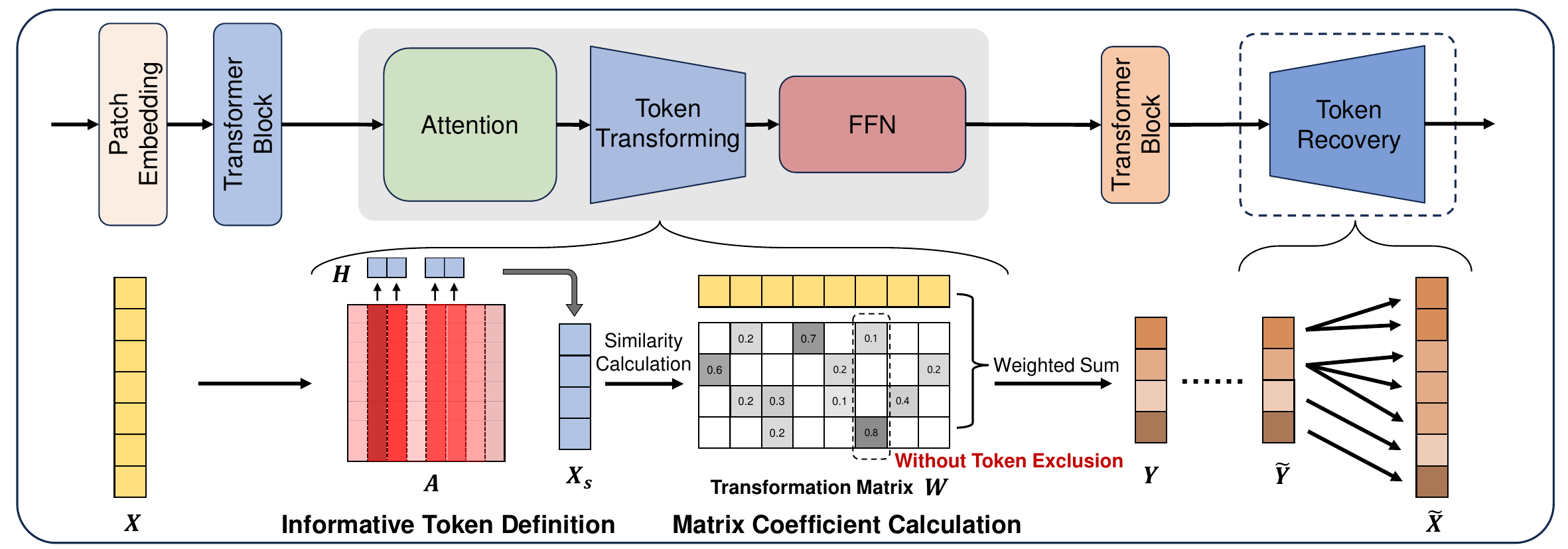}
\vspace{-5pt}
\caption{Detail structure of our proposed Token Transforming framework.  The module is inserted between the self-attention layer and FFN layer. Our framework dynamically selects the most informative tokens based on the attention map, calculates the matrix coefficient through similarity calculation of informative tokens, and finally obtains a reduced set of transformed tokens via a weighted sum without the constraint of token exclusion. The token recovery module is applied to dense prediction tasks to obtain dense tokens.}
\vspace{-5pt}
\label{fig:detail-structure}
\end{figure*}

As shown in~\cref{fig:detail-structure}, we apply Token Transforming module between the self-attention layer and FFN layer to reduce the number of tokens. The determination of transformation matrix is divided into two key steps: \textbf{informative token definition} and \textbf{matrix coefficient calculation}. We first select $M$ as the most informative token from the original $N$ tokens. Then, we calculate the gate similarity between these informative tokens and full $N$ tokens, which reflects weighted importance coefficient. The normalized gate similarity values are then used to represent the coefficient in $\boldsymbol{W}$. Finally, $M$ transformed tokens are calculated by transformation along with attention scaling adaptation for efficient inference. For dense prediction tasks, we add an additional Token Recovery module to reconstruct dense tokens illustrated in~\cref{sec:adapt_dense_prediction}.

\subsection{Informative Token Definition}
\label{sec:token_definition}
Informative tokens are determined by token selection criterion to obtain primary queries for matrix. 
Previous methods~\cite{rao2021dynamicvit, liang2022not, wei2023joint} use the attention map between the class token and other tokens as the token selection criterion, limiting their applicability to image classification tasks only. In contrast, we compute the informativeness level of each token directly from the attention map:
\begin{equation}
    \boldsymbol{H_j} = \sum_{i=1}^{N}{\boldsymbol{A_{ij}}}, \quad j=1,\cdots,N,
\end{equation}
where $\boldsymbol{A_{ij}}$ is the element in the $i^{th}$ row and $j^{th}$ column of attention map. Intuitively, $\boldsymbol{H}$ represents the informativeness level of tokens. 
A higher $\boldsymbol{H_j}$ indicates that a token provides more information to others. We then select the $M$ most informative tokens $\boldsymbol{X}_s$ to construct subset $\mathbb{D}_{s}$ from full token set $\mathbb{D}$ based on sorting. Notably, our token selection process is independent of the class token, enabling the method to effectively handle dense prediction tasks.

\subsection{Matrix Coefficient Calculation}
\label{sec:token_updating}
\textbf{Gate similarity.}
Once the informative tokens are defined, we use them as queries to calculate similarity, measure weighted importance of full tokens and reserve global spatial information. Concretely, we calculate cosine similarity between informative tokens and full tokens. Since unrelated tokens may introduce noise and disrupt token transformation in a global context, we mitigate this by applying a threshold $\kappa$ to filter out irrelevant tokens: 

\begin{equation}
    \mathrm{GSim}(i,j)= \left \{
     \begin{array}{cc}
        \mathrm{sim}(i,j),& \mathrm{sim}(i,j) \geq \kappa \\
        0,&  \mathrm{sim}(i,j) < \kappa \\
    \end{array},
    \right.
\end{equation}
and cosine similarity is defined as:
\begin{equation}
    \mathrm{sim}(i,j)=\frac{\boldsymbol{X}_i\cdot\boldsymbol{X}_j}{\left\| \boldsymbol{X}_i \right\| \left\| \boldsymbol{X}_j \right\|},
\end{equation}
where $\boldsymbol{X}_i\in\mathbb{D}_{s}$ and $\boldsymbol{X}_j\in\mathbb{D}$. This ensures that only the most relevant tokens are associated with each informative token. 
Intuitively, compared with methods using local aggregation size~\cite{liang2022expediting, bolya2022token}, the attachment to full tokens reserves information to the most in a denoised and global spatial aggregation size. 

\noindent \textbf{Coefficient normalization.}
The weighted coefficient of $\boldsymbol{W}$ determines the degree to which full tokens contribute to informative tokens.
Since each token $\boldsymbol{X}_j$ may be assigned to more than one transformed token in a non-exclusive manner, it is crucial to convert absolute coefficient into relative ones. To achieve this,
we introduce assignment normalization based on $\rm Softmax$ operation with temperature, ensuring that the coefficients are properly scaled:
\begin{equation}
    m_{ij} = \frac{{\rm exp}\Big({\mathrm{GSim}(i,j)*\tau}\Big)}{\sum_{k=1}^M{{\rm exp}\Big(\mathrm{GSim}(k,j)*\tau}\Big)},
\end{equation}
Here, $\mathrm{GSim}(i,j)$ represents the gated similarity score between the $i$-th informative token and the $j$-th full token, while $\tau$ is a temperature parameter that controls the sharpness of the distribution.  This assignment normalization is performed along the column dimension of $\mathrm{GSim}(i,j)$, as illustrated in~\cref{fig:detail-structure}. This distinguishes our approach from existing token-exclusive methods, where each column contains at most one nonzero value, restricting token assignment flexibility. The following experiments demonstrate the effectiveness of assignment normalization. Notably, omitting this step leads to a severe accuracy drop.

Besides assignment normalization, we apply row-wise normalization to ensure that the final weighted coefficients sum to one across each row. This guarantees that the transformed tokens remain a proper weighted sum of the full tokens. After obtaining the final transformation matrix $W$, the transformed token $\boldsymbol{Y}$ are obtained as follows:
\begin{equation}
W_{ij} = \frac{m_{ij}}{\sum_{j=1}^N{m_{ij}}},\quad
    {\boldsymbol{Y}_i} = \sum_{j=1}^N{W_{ij}}{\boldsymbol{X}_{j}},
\end{equation}
where $W_{ij}$ represents the normalized weight assigned to token $\boldsymbol{X}_j$. The transformed token $\boldsymbol{Y}_i$ participates in the following blocks for training-free acceleration. Notably, our method benefits from efficient matrix multiplication implementations, leading to faster inference, as shown  in~\cref{sec:results-classification}.

\subsection{Attention Scaling Adaptation}
\label{sec:scaling}
As tokens aggregate information to varying degrees during transformation, the transformed tokens no longer have uniform quantities. It is inappropriate to use standard self-attention to process these tokens. Therefore, we add scaling adaptation to adjust the attention map accordingly:
\begin{equation}
    \boldsymbol{A} = {\rm Softmax}\Big(\frac{{\boldsymbol{Q}\boldsymbol{K}^T}}{\sqrt{d}} +{\rm log}\mathbf{{\rm (broadcast}(\boldsymbol{s}))}\Big),
\label{eqn:attn-def}
\end{equation}
where $\rm broadcast(\cdot)$ stands for repeating row vector along column axis and $\boldsymbol{s}$ is a row vector with length of $M$. Conceptually, $\boldsymbol{s}$ represents the quantity of tokens and adjusts the $\rm Softmax$ output by increasing the attention weights of tokens with large quantity. Thus, $\boldsymbol{s}$ serves as a scaling adaptation factor to adjust attention weights, aligning them with token importance. The scaling adaptation $\boldsymbol{s}$ for each transforming token is obtained from relative coefficient:
\begin{equation}
    s_i = \sum_{j=1}^N{m_{ij}},\ i=1,\cdots,M,\ j=1,\cdots,N.
\end{equation}
Note that our scaling adaptation $\boldsymbol{s}$ in~\cref{eqn:attn-def} differs fundamentally from that in ToMe~\cite{bolya2022token}, which defines token size as an integer representing actual token count and is unsuitable for many-to-many paradigm. In contrast, our scaling adaptation is a continuous value derived from gate similarity,  representing a soft token assignment relationship rather than a discrete token count. 

\subsection{Extension to Dense Prediction Tasks}
\label{sec:adapt_dense_prediction}
Since the token transforming process reduces the number of tokens, dense prediction tasks, which require full-resolution token outputs, necessitate an additional token recovery step. Given that transformed tokens effectively capture rich and generalizable representations, we employ a simple nearest neighbor method for token recovery. Specifically, we first compute the distance from each original token $\boldsymbol{X}$ to all transformed tokens $\boldsymbol{Y}$ using $L_2$ norm, respectively. 
Then, we record the nearest neighbor index $I$ for each original token as follows:
\begin{equation}
    I_m = \mathop {\arg \min }\limits_i{\left\| {{\boldsymbol{X}_m} - {\boldsymbol{Y}_i}} \right\|}, m=1,\cdots,N.
\end{equation}

The dense tokens can therefore be reconstructed in the recovery layer using the recorded index relation as follows:
\begin{equation}
    \boldsymbol{\widetilde{X}}_{m} = \boldsymbol{\widetilde{Y}}_{I_m},\quad m=1,\cdots,N,
\end{equation}
where $\boldsymbol{\widetilde{Y}}$ denotes the tokens before recovery layer and $\boldsymbol{\widetilde{X}}$ denotes fully reconstructed dense tokens. The reconstructed tokens can be seamlessly integrated into task-specific prediction heads, allowing for effective adaptation to segmentation, object detection, depth estimation, and other dense prediction tasks.

\section{Experiment}
We conduct various experiments and ablation study on image classification to demonstrate the effectiveness of our method. Due to its non-parametric property and decoupling with class token, we further extend our approach as a plug-and-play module to dense prediction tasks including depth estimation, semantic segmentation, object detection and large language model to prove the transferability and scalability of our method.

\subsection{Experimental Setup}
\noindent \textbf{Datasets.} We use ImageNet~\cite{deng2009imagenet} for image classification, ADE20k~\cite{zhou2019semantic} and Cityscapes~\cite{cordts2016cityscapes} for semantic segmentation, COCO~\cite{lin2014coco} for object detection, NYUv2~\cite{silberman2012indoor} for depth estimation, and finally, TextVQA~\cite{singh2019towards}, VQAv2~\cite{goyal2017making}, ScienceQA~\cite{lu2022learn}, POPE~\cite{li2023evaluating} for language model generation.

\noindent \textbf{Evaluation Metrics and Implementation Details.} For image classification, we employ our method at certain stages of transformer layers. We report Top-1 accuracy for performance comparison and throughput for efficiency measurement. We use $``/"$ after model name to indicate the reserving ratio of each transforming layer. For example, $``/$0.7$"$ means reserving 70\% of tokens after each transforming. We set reserving ratio to 0.7 by default unless otherwise stated.

For dense prediction tasks, depth estimation is primarily assessed by AbsRel, RMSE and we report additional metrics listed in DPT~\cite{ranftl2021vision}. In segmentation, we incorporate mIoU as the metric and evaluate Segmenter-L on both ADE20k and Cityscapes. For object detection, we choose state-of-the-art object detection network DINO with Swin-L backbone, and report mAP as measurement. For large model generation, we report response accuracy on commonly evaluated datasets. 

\subsection{Main Results on Image Classification}
\label{sec:results-classification}
We conduct extensive experiments on different ViTs and compare with existing methods to showcase the superiority, efficiency, scalability and generalizability of our method.

\begin{figure*}
    \centering
    \includegraphics[width=\linewidth]{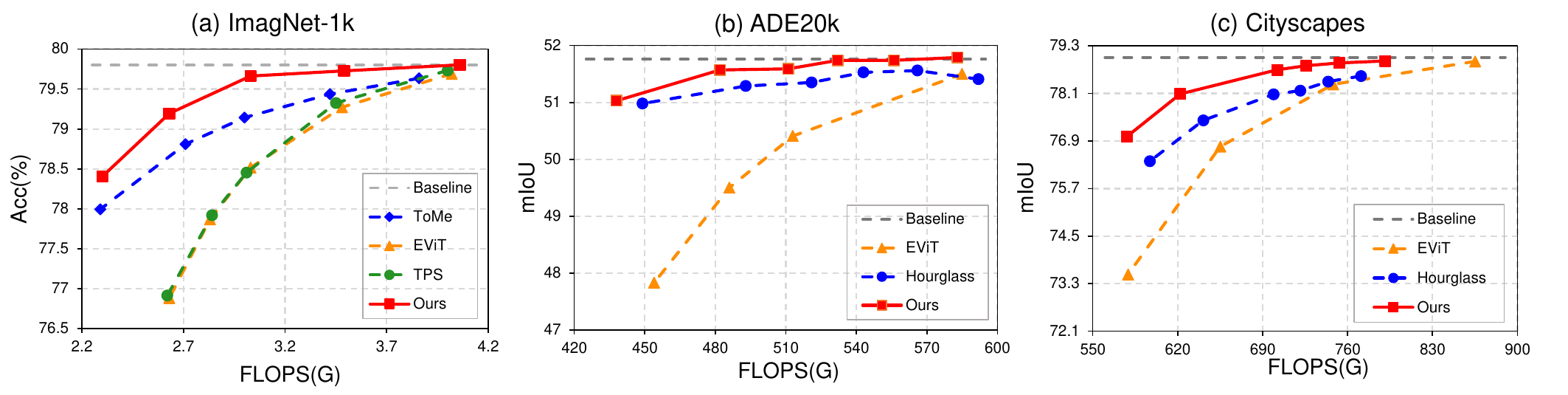}
    \vspace{-20pt}
    \caption{Comparison with previous methods under different FLOPs in a training-free manner on different tasks. For fair comparison, we keep the same token reduction configurations as other methods. (a) ImagNet-1k for image classification (b) ADE20K (c) Cityscapes for segmentation.  More Results are obtained by running official codes due to limited data in papers. Our method gains consistent and significant improvements under different settings especially under aggressive compression ratio.}
    \vspace{-10pt}
    \label{fig:main-curve}
\end{figure*}

\noindent \textbf{Training-free Results.} We compare our approach with other methods on DeiT~\cite{touvron2021training}.  We first insert token transforming as a plug-and-play plugin into transformer layers in a training-free manner. As is shown in~\cref{tab:main-results}, we achieve superior performance. For example, we compress the model by up to 34.8\% with marginally loss in accuracy on DeiT-S. Moreover, the accuracy of Deit-S compression result on the fly is significantly higher than other methods~(\textbf{+0.2}). We also evaluate our method of different compression ratio and draw accuracy-FLOPs curves compared with other methods~\cite{bolya2022token,liang2022not,wei2023joint} under same off-the-shelf setting. From~\cref{fig:main-curve}\textcolor{arxivblue}{a}, our method achieves consistent and significant improvements especially towards aggressive compression and obtains \textbf{lossless compression} of 30\% in a training-free manner, demonstrating the usefulness of our approach.

\begin{table}[t]
\renewcommand{\arraystretch}{1.25}
\centering
\caption{Comparison of various dynamic compression on ViTs.}
\setlength{\tabcolsep}{4pt}
    \vspace{-10pt}
    \resizebox{\linewidth}{!}{
    \Large
    \begin{tabular}{lccccc}
    \toprule[1.3pt]
    Model &Training-Free& Params~(M) & GFLOPs & Acc~(\%) & Throughput~(im/s)\\ \midrule
    \rowcolor[gray]{0.9} DeiT-S~\cite{touvron2021training} &-& 22.1 & 4.6 & 79.8 &974  \\
    DynamicViT~\cite{rao2021dynamicvit} &\ding{55} &22.8 & 3.0~(34.8\% $\downarrow$) & 79.3  &1503 \\
    Evo-ViT~\cite{xu2022evo} & \ding{55} & 22.4 & 3.0~(34.8\% $\downarrow$) & 79.4 &1510 \\
    EViT~\cite{liang2022not} & \ding{55} &22.1 & 3.0~(34.8\% $\downarrow$) & 79.5 & 1487\\
    ATS~\cite{fayyaz2022adaptive} & \ding{55} &22.1 & 2.9~(37.0\% $\downarrow$) & 79.7 & 1382 \\
    ToMe~\cite{bolya2022token} & \ding{55} &22.1 & 2.7~(41.3\% $\downarrow$) & 79.4 & 1552 \\   
    TPS~\cite{wei2023joint}&\ding{55} &22.1 & 3.0~(34.8\% $\downarrow$) & 79.7 &1428 \\
    \rowcolor[rgb] {1,1, 0.848} \bftab{Ours/\bftab{0.7}} &\ding{55} & 22.1 & 3.0~(34.8\% $\downarrow$) & \bftab{79.9}~\textcolor{Maroon}{(+0.2)} &1451\\
    \rowcolor[rgb] {1,1, 0.848} \bftab{Ours/\bftab{0.6}} &\ding{55} & 22.1 & \bftab{2.6}~(43.5\% $\downarrow$) & \bftab{79.7}&\bftab{1633}\\
    \rowcolor[rgb] {1,1, 0.848} \bftab{Ours} & $\checkmark$&22.1 & 3.0~(34.8\% $\downarrow$) & \bftab{79.7} &1451\\
    \midrule
    \rowcolor[gray]{0.9} ViT-Augreg-S~\cite{steiner2022train} &-& 22.1 & 4.6 & 81.4  &$974$ \\
    ToMe~\cite{bolya2022token} &$\checkmark$& 22.1 & 2.7~(41.3\% $\downarrow$) & 79.3 &1564\\   
    \bftab{Ours} &$\checkmark$& 22.1 & 2.7~(41.3\% $\downarrow$) & \bftab{79.8}~\textcolor{Maroon}{(+0.5)}&\bftab{1576}\\
    \midrule
    \rowcolor[gray]{0.9} ViT-AugReg-Ti~\cite{steiner2022train}&-& 5.6 & 1.3 & 75.5 &2558\\
    ToMe~\cite{bolya2022token} & $\checkmark$&5.6 & 0.8~(38.5\% $\downarrow$) & 73.8& 3629\\ 
    \rowcolor[rgb] {1,1, 0.848} \bftab{Ours} & $\checkmark$&5.6 & 0.8~(38.5\% $\downarrow$) & \bftab{74.6}~\textcolor{Maroon}{(+0.8)}& \bftab{3639}\\
    \midrule
    \rowcolor[gray]{0.9} ViT-AugReg-B~\cite{steiner2022train}&- & 86.6 & 17.6 & 84.5&309 \\
    ToMe~\cite{bolya2022token}&$\checkmark$& 86.6 & 11.6~(34.1\% $\downarrow$) & 83.3& 464\\
    \rowcolor[rgb] {1,1, 0.848} \bftab{Ours} & $\checkmark$&86.6 & \bftab{11.4}~(\bftab{35.2}\% $\downarrow$) & \bftab{83.7}~\textcolor{Maroon}{(+0.4)}&\bftab{469} \\
    \midrule
    \rowcolor[gray]{0.9} ViT-H~\cite{he2022masked} &- & 632.1 & 167.4 & 86.9 &35\\
    ToMe~\cite{bolya2022token}& \ding{55}&632.1 & 92.9~(44.5\% $\downarrow$) & 86.5 &63\\
    \rowcolor[rgb] {1,1, 0.848} \bftab{Ours} &\ding{55}& 632.1 & 92.9~(44.5\% $\downarrow$) & \bftab{86.7}~\textcolor{Maroon}{(+0.2)}&\bftab{66}\\
    ToMe~\cite{bolya2022token}&$\checkmark$& 632.1 & 92.9~(44.5\% $\downarrow$) & 85.9&63\\
    \rowcolor[rgb] {1,1, 0.848} \bftab{Ours}&$\checkmark$& 632.1 & 92.9~(44.5\% $\downarrow$) & \bftab{86.1}~\textcolor{Maroon}{(+0.2)}&\bftab{66}\\
\bottomrule[1.3pt]
\end{tabular}}
\vspace{-15pt}
\label{tab:main-results}
\end{table}

\begin{table}[ht]
\centering
\setlength{\tabcolsep}{6pt}
\renewcommand{\arraystretch}{1.25}
\Large
\caption{{Comparison with other ViT structure based model.}}
    \vspace{-10pt}
    \resizebox{\linewidth}{!}
    {
    \begin{tabular}{lcccc}
    \toprule[1.3pt]
    Model &Training-Free& Params~(M) & GFLOPs & Acc~(\%)  \\ 
    \midrule
    \rowcolor[gray]{0.9} T2T-ViT-14~\cite{yuan2021tokens} &-&21.5&4.8 & 81.5\\
    PS-T2T-14~\cite{tang2022patch} &\ding{55}& -  & 3.1 & 81.3 \\
    \rowcolor[rgb] {1,1, 0.848} \bftab{Ours-T2T-}\bftab{14}&\ding{55}& 21.5 & 3.1& \bftab{81.5}~\textcolor{Maroon}{(+0.2)}\\
    \rowcolor[rgb] {1,1, 0.848} \bftab{Ours-T2T-}\bftab{14}&$\checkmark$& 21.5 & 3.1& \bftab{81.3}\\
    \midrule
    
    \rowcolor[gray]{0.9} PS-ViT-B~\cite{yue2021vision} &-& 21.3 & 5.4 & 81.7\\
    ATS-PS-B~\cite{fayyaz2022adaptive}& \ding{55}&21.3 & 3.7 & 81.5\\
    \rowcolor[rgb] {1,1, 0.848} \bftab{Ours-PS-B}& \ding{55}&21.3 & 3.7 & \bftab{81.7}~\textcolor{Maroon}{(+0.2)} \\
    \rowcolor[rgb] {1,1, 0.848} \bftab{Ours-PS-B}&$\checkmark$& 21.3 & 3.7 & \bftab{81.5}\\
    \midrule
    \rowcolor[gray]{0.9} LV-ViT-S~\cite{jiang2021all} &-& 26.2 & 6.6 & 83.3 \\
    DynamicViT-LV-S~\cite{rao2021dynamicvit} &\ding{55}& 26.9 & 4.6 & 83.0  \\
    PS-LV-ViT-S~\cite{tang2022patch}&\ding{55} & 26.2 & 4.7 & 82.4 \\
    EViT-LV-S~\cite{liang2022not}&\ding{55}&26.2 & 4.7 & 83.0\\
    \rowcolor[rgb] {1,1, 0.848} \bftab{Ours-LV-S}&\ding{55}& 26.2 & 4.6 & \bftab{83.3}~\textcolor{Maroon}{(+0.3)} \\
    \rowcolor[rgb] {1,1, 0.848} \bftab{Ours-LV-S}&$\checkmark$& 26.2 & 4.6 & \bftab{83.1}~\textcolor{Maroon}{(+0.1)} \\
\bottomrule[1.3pt]
\label{tab:compare_lv_vit}
\end{tabular}}
\vspace{-25pt}
\end{table}

\noindent \textbf{Scalability to larger ViTs.} We carry out experiments on larger MAE pre-trained ViT-H~\cite{he2022masked} and ViT-AugReg~\cite{steiner2022train} to evaluate scalability of our method. \Cref{tab:main-results} strongly certificates that our method scales well and gets significant 0.2\%, 0.5\%, 0.8\% and 0.6\% improvement with faster inference.

\noindent \textbf{Inference Acceleration.} We report throughput on V100 to substantiate inference acceleration. Outcomes in~\cref{tab:main-results} showcase that we accelerate DeiT-S by $\times1.5$ and achieve consistent higher throughput on all methods. We highlight that the improvement in acceleration can be attributed to efficient matrix multiplication implementation of token reduction process, which is faster than token-wise mapping and reduction in existing methods.

\noindent \textbf{Robustness to different variants of ViTs.} To further validate the generalizability of our method, we compare our method with other ViT based models that achieve progress on image classification. Specifically, we apply our method on LV-ViT, PS-ViT and T2T-ViT. As shown in~\cref{tab:compare_lv_vit}, it turns out that our compression results of LV-ViT-S without fine-tuning outperforms the previous fine-tuning methods by 0.1\% accuracy promotion under comparable compression ratio, which indicates the significant improvement. 
We also implement our method to PS-ViT~\cite{yue2021vision} and T2T-ViT~\cite{yuan2021tokens} and it reveals in~\cref{tab:compare_lv_vit} that our proposed method can achieve competitive results against previous methods. 

\noindent \textbf{Further tuning improves the performance.} Although the proposed approach can accelerate ViTs in a training-free manner to some degree, fine-tuning is also beneficial to improving the performance. Considering most methods only report fine-tuned results, we also give fine-tuned results for 30 epochs, which is a common training configuration, to make fair comparison. We report fine-tuning results in~\cref{tab:main-results}, where our approach consistently outperforms all mentioned methods with state-of-the-art results. Specifically, our model even gains +0.1\%~(79.9\%) accuracy bonus \textbf{under standard compression} and get merely -0.1\%~(79.7\%) accuracy drop \textbf{towards more aggressive compression}~(2.6 GFLOPs) compared to vanilla DeiT-S, consistently outperform previous works by a solid margin. Moreover, fine-tuned ViT-H obtains better results than ToMe~(86.7\%v.s. 86.5\%) with considerable 45\% compression. Also, compression results of all three variants in~\cref{tab:compare_lv_vit} achieve \textbf{no accuracy drop} after fine-tuning and outperform the existing methods by more than +0.2\%. All results strongly validate the effectiveness of our method.

\begin{table*}[ht]
\setlength{\tabcolsep}{4pt}
\renewcommand{\arraystretch}{1.25}
	\centering
         \caption{Comparison with Hourglass on depth estimation when applying DPT as baseline.}
         \vspace{-8pt}
 \resizebox{\linewidth}{!}{
	\begin{tabular}{lcccccccccc}
		\toprule[1.3pt]
		  Method  & \small{$\delta>1.25$}~$\uparrow$ & \small{$\delta>1.25^2$}~$\uparrow$ & \small{$\delta>1.25^3$}~$\uparrow$ & AbsRel~$\downarrow$ & SqRel$\downarrow$ & RMSE$\downarrow$ & RMSElog~$\downarrow$ & SILog~$\downarrow$ & log10~$\downarrow$ & GFLOPs~$\downarrow$\\ 
            \midrule  
        \rowcolor[gray]{0.9} DPT~\cite{ranftl2021vision} & 0.904 &0.988 & 0.998 & 0.110 & 0.054 & 0.357 & 0.129 & 9.524 & 0.045 & 280 \\
        DPT+Hourglass~\cite{liang2022expediting} & 0.900 & 0.987 & 0.998 & 0.113  & 0.056 & 0.363 & 0.132 & \bftab{9.532} & 0.046 & 202 \\
        \rowcolor[rgb] {1,1, 0.848} \bftab{DPT+Ours} &  
        \bftab{0.901} & \bftab{0.988} & \bftab{0.998} & \bftab{0.108} & \bftab{0.054} & \bftab{0.362} & \bftab{0.130} & 9.536 & \bftab{0.045} & \bftab{200} \\
        \bottomrule[1.3pt]
	\end{tabular}}
	\label{fig:dpt_results}
\vspace{-12pt}
\end{table*}

\subsection{Semantic Segmentation}
\noindent \textbf{Settings.} We conduct additional experiments on semantic segmentation, select Segmenter-L/16 as the baseline. Specifically, we apply our method in Vision Transformer block of Segmenter and compare our method with Hourglass~\cite{liang2022expediting} and EViT~\cite{liang2022not} in a training-free manner to have a fair comparison with the two methods.

\noindent \textbf{Results.} We conduct serveral experiments under different computational cost.
As shown in~\cref{fig:main-curve}\textcolor{arxivblue}{b} and~\cref{fig:main-curve}\textcolor{arxivblue}{c}, the results  demonstrate that our method can achieve minimal mIoU drop under comparable compression of FLOPs on different datasets, especially that the lower compressed FLOPs is, the more advantage our method has. For example, on ADE20k, our framework obtains \textbf{30\% lossless compression} and on Cityscapes, our method significantly outperforms Hourglass by 0.4\%~(78.8 v.s. 78.4) under the same compression settings, with notable \textbf{25\% lossless compression}. 
We highlight that capability of reserving token information to the most under many-to-many framework contributes crucially to superior performance in complex dense prediction tasks, which is validated by sub-optimal results of token-exclusive reduction methods like EViT. 

\subsection{Monocular Depth Estimation}
\noindent \textbf{Settings.} To further demonstrate generalization of the proposed method, we implement our method in DPT~\cite{ranftl2021vision} built on hybrid vision transformer ResNet50+ViT-B/16, where a ResNet-50 feature extractor is applied for following ViT. 

\noindent \textbf{Results.} We report comprehensive evaluation on NYUv2 in~\cref{fig:dpt_results} and results indicate that our method is capable of accelerating FLOPs by 29\% with AbsRel even lower than original DPT~(0.108 v.s. 0.110). Moreover, our method outperforms Hourglass~\cite{liang2022expediting} in most of the metric with lower computational cost, deeply showcasing the superiority.

\subsection{Object Detection}
\noindent \textbf{Settings.} We carry out experiments on object detection, in which we incorporate DINO~\cite{zhang2022dino} with Swin-L~\cite{liu2021swin} backbone. As class token not longer exists, most dynamic compression methods which use metrics related to class attention map and are designed for classification task cannot be applied any more. Hence, we compare with Hourglass~\cite{liang2022expediting} under same compression settings in following experiments.

\begin{table}[h]
    \centering
    \caption{Detailed comparison for object detection on COCO.}
    \vspace{-8pt}
    \resizebox{\linewidth}{!}{
    \begin{tabular}{lcccc}
    \toprule[1.3pt]
    Model &Backbone & GFLOPs$\downarrow$ & FPS$\uparrow$ & mAP$\uparrow$\\
    \midrule
    \rowcolor[gray]{0.9} DINO~\cite{zhang2022dino}&Swin-L & 936 &4.74 &58.5 \\
    \quad +Hourglass~\cite{liang2022expediting} &Swin-L& 746 & 5.64 & 56.9\\
    \rowcolor[rgb] {1,1, 0.848} \bftab{\quad +Ours} & Swin-L& \bftab{734}& \bftab{5.72} & \bftab{57.3}~\textcolor{Maroon}{(+0.4)} \\
    \bottomrule[1.3pt]
    \end{tabular}}
    \vspace{-8pt}
    \label{tab:compare_detection}
\end{table}

\noindent \textbf{Results.} We compare our proposed framework with existing method under similar computation and the results are shown in~\cref{tab:compare_detection}. It indicates that firstly, our method can achieve better performance against Hourglass with notable 0.4\% higher mAP, higher FPS and less computation, which can be attributed to efficient and effective matrix implementation. Moreover, the results show that the proposed method is also effective in larger visual tasks like object detection and is compatible with Swin-like backbone with satisfying results, which strongly verifies the transferability and versatility of the method.

\begin{table}[h]
    \centering
    \setlength\tabcolsep{3pt}
    \renewcommand{\arraystretch}{1.2}
    \caption{Comparison results on multiomdal large language models by accelerating visual encoder.}
    \vspace{-10pt}
    \footnotesize
    \resizebox{\linewidth}{!}{
    \begin{tabular}{l>{\centering\arraybackslash}p{2cm} >{\centering\arraybackslash}p{1cm} >{\centering\arraybackslash}p{2cm}>{\centering\arraybackslash}p{1cm}}
    \toprule[1.3pt]
         Method & ScienceQA & TextVQA & VQAv2 & POPE \\
         \midrule
       \rowcolor[gray]{0.9} LLaVA~\cite{liu2024visual}  & 68.4 &58.2 & 79.1 &86.4\\
       \quad +ToMe~\cite{bolya2022token}& 50.0&45.3&57.1&52.5\\ 
       \quad +PruMerge~\cite{shang2024prumerge} & 68.5 &  53.5& 65.9&70.7\\
       \rowcolor[rgb] {1,1, 0.848} \quad +\bftab{Ours} & \bftab{69.3} & \bftab{54.6} & \bftab{72.5} &\bftab{71.6} \\
    \bottomrule[1.3pt]
    \end{tabular}}
    \vspace{-10pt}
    \label{tab:results-mllms}
\end{table}

\subsection{Multimodal Large Language Models}
\noindent \textbf{Settings.} We further accelerate multimodal large language model~\cite{zeng2025parameter} by compressing visual encoder of LLaVA~\cite{liu2024visual}, which generates response auto-regressively, using LLaVA-v1.5-7b as base model and compare our method with ToMe~\cite{bolya2022token}, PruMerge~\cite{shang2024prumerge} under comparable compression ratio to examine the effectiveness in modern large models.

\noindent \textbf{Results.} It is shown in~\cref{tab:results-mllms} that our method achieves the best results on both retaining task-specific knowledge~(ScienceQA, TextVQA, VQAv2) and maintaining comprehensive skills like detecting hallucination~(POPE), substantially outperforming existing methods by a solid margin. The results strongly indicate the utility of our method in accelerating modern multimodal large language models.

\begin{table}[b]
    \centering
        \renewcommand{\arraystretch}{1.25}
        \setlength{\tabcolsep}{6pt}
        \vspace{-10pt}
        \caption{Influence of hyper-parameter selection.}
        \vspace{-8pt}
        \footnotesize
        \resizebox{\linewidth}{!}{
        \begin{tabular}{ c | cccccc}
        \toprule[1.3pt]
        {Threshold $\kappa$} & 0.3 & 0.4 & 0.5 & 0.6 & 0.7 & 0.8\\
        \midrule
        Acc~(\%) & 79.1 & 79.5 & \bftab{79.7} & 79.6 & 79.5 & 78.3 \\
        \midrule
        \midrule
        {Temperature $\tau$} & 1& 20  & 100 & 150 & 200 & 250 \\
        \midrule
        Acc~(\%) & 74.5 & 79.2 & 79.5 & \bftab{79.7} & 79.5& 79.4 \\
        \bottomrule[1.3pt]
        \end{tabular}}
        \vspace{-10pt}
        \label{tab:hyperparameter}
\end{table}

\subsection{Ablation Study}
We conduct comprehensive ablation study on image classification with DeiT-S to verify the effectiveness of each component and influence of different hyper-parameters. 

\noindent \textbf{Hyper-parameter selection of token transforming.} We begin with different hyper-parameters related to our framework. Concretely, we study influence of gate similarity threshold $\kappa$ and temperature $\tau$, respectively. As shown in~\cref{tab:hyperparameter}, performance of the framework threshold $\kappa$ achieves the best result between 0.5 and 0.6, which is consistent with the analysis that lower threshold brings in noisy disturbance and higher threshold would block important tokens. For temperature, the performance becomes better as temperature initially increases and levels off over a wide range as temperature varies, which verifies the robustness of our approach.

\begin{table}[t]
    \centering
    \caption{\footnotesize Influence of scaling adaptation and assignment normalization.}
    \vspace{-8pt}
    \setlength{\tabcolsep}{8pt}
    \renewcommand{\arraystretch}{1.15}
    \resizebox{\linewidth}{!}{
    \begin{tabular}{c|cccc}
    \toprule[1.3pt]
        Method  & Scaling & Normalization  & GFLOPs & Acc~(\%) \\
        \midrule
        DeiT-S&&&4.6&79.8 \\ 
        \midrule
        \multirow{3}{*}{DeiT-S/0.7}&\checkmark  &  & 3.0 & 5.7 \\
         &  & \checkmark & 3.0 & 77.9 \\
        &\checkmark & \checkmark & 3.0 & \bftab{79.7} \\
        \bottomrule[1.3pt]
    \end{tabular}}
    \vspace{-3pt}
    \label{tab:scaling-adaptation}
\end{table}

\begin{table}[t]
    \centering
    \vspace{-3pt}
    \caption{Influence of different token selection criterion and similarity score in transformation matrix construction.}
    \footnotesize
    \vspace{-8pt}
    \resizebox{\linewidth}{!}{
    \begin{tabular}{lcc}
    \toprule[1.3pt]
        Strategy & GFLOPs & Acc~(\%) \\
        \midrule
        DeiT-S&4.6&79.8 \\ 
        \midrule
       \multicolumn{3}{c}{{Token selection criterion}} \\
        \midrule
         Uniform & 3.0 & 78.3\\
         Class token attention & 3.0 & 79.5\\
         \rowcolor[rgb] {1,1, 0.848} Informative token attention~\bftab{(Ours)}& 3.0 & \bftab{79.7}\\
        \midrule
        \multicolumn{3}{c}{{Similarity score}} \\
        \midrule
        Euclidean distance & 3.0 & 78.5\\
        \rowcolor[rgb] {1,1, 0.848} Cosine distance~\bftab{(Ours)}& 3.0 & \bftab{79.7}\\
        \bottomrule[1.3pt]
    \end{tabular}}
    \vspace{-15pt}
    \label{tab:ablation-strategy}
    \end{table}

\noindent \textbf{Effectiveness of scaling adaptation.} Scaling adaptation represents the float quantity of tokens after transformation. \Cref{tab:scaling-adaptation} shows its significance that framework with scaling adaptation leads to impressive gain in accuracy~(77.9\% v.s. 79.6\%). It strongly demonstrates the necessity and rationality of increasing attention weights of tokens with large quantity to adjust $\rm Softmax$ output under general many-to-many framework for competitive performance.

\noindent \textbf{Influence of assignment normalization.} Assignment normalization plays a vital role in the effectiveness of the framework as flexibility non-exclusive property requires proper coefficient conversion of each column in matrix. We show in ~\cref{tab:scaling-adaptation} that framework without assignment normalization can lead to extremely low accuracy under standard setting~(5.7\% v.s. 79.6\%). A possible reason for accuracy collapse is that original absolute coefficient fails to build token connection between many-to-many paradigm without assignment normalization. Thus, the importance of assignment normalization is deeply verified.

\noindent \textbf{Comparison with different token selection criterion and similarity strategy.}  We analyze the necessity of each component proposed in our framework, \ie token selection criterion and similarity score for matrix construction. For the former, we compare our informative token criterion with uniform and class attention selection, which refer to defining the important tokens according to adaptive average pool~\cite{liang2022expediting} and attention with class token~\cite{liang2022not}, respectively. For the latter, we replace our cosine distance based score with euclidean distance one. As shown in~\cref{tab:ablation-strategy} , no matter what the alternative approach is, there is a sharp accuracy drop except class token attention. However, the token selection criterion related to class token cannot be extended to dense prediction tasks. Thus, the effectiveness and transferability of each component is demonstrated.

\section{Visualization}
\label{sec:visual}

\begin{figure}[t]
  \centering
    \includegraphics[width=\linewidth]{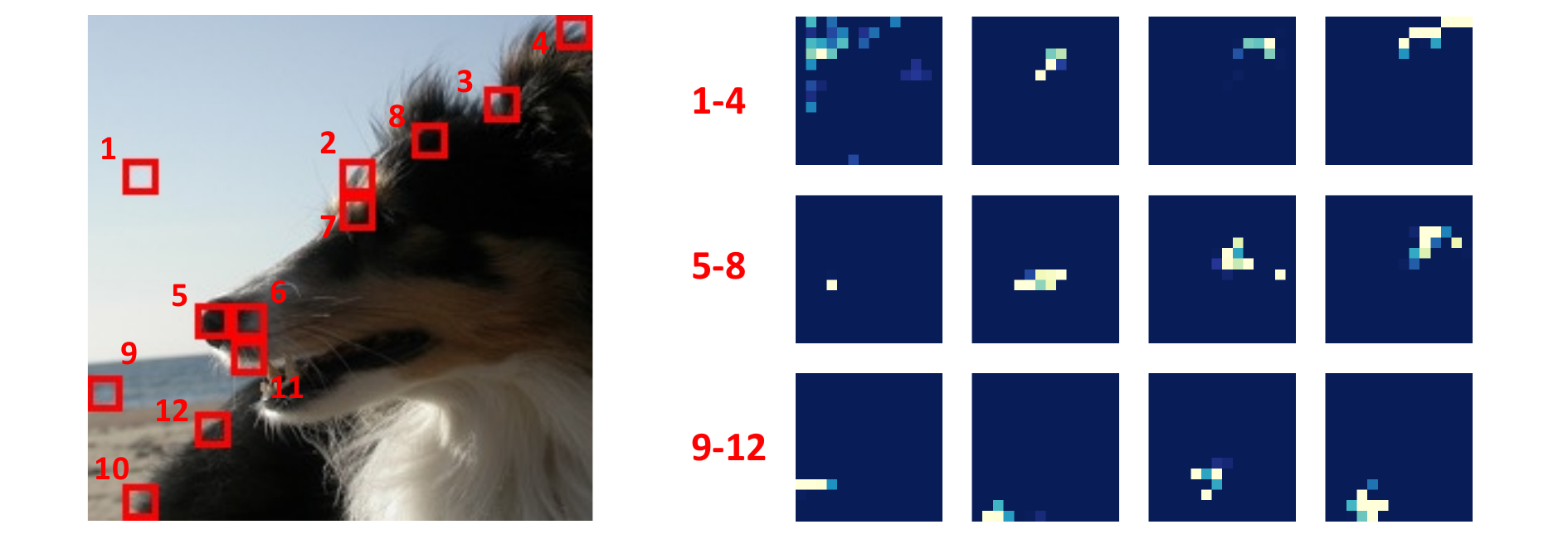}
    \vspace{-20pt}
\caption{Informative tokens and the heatmap of transformation matrix coefficient for each informative token. Lighter color represents greater coefficient value. One original token can be assigned to multiple informative tokens, reflecting non-exclusive property. }
   \label{fig:visualization}
   \vspace{-15pt}
\end{figure}

One key exploration of our transformation matrix is non-exclusive property of token assignment. It means that original tokens can be integrated into more than one crucial remaining token after token reduction, which previous methods can not. To have an intuitive understanding of the property, we provide heatmap for several typical informative tokens with respect to their transformation matrix coefficient. In~\cref{fig:visualization}, informative tokens tagged with 2, 3 and 8 share some common assigned tokens, as corresponding heatmap have overlapped high activation area and therefore repeatedly aggregating information from several crucial tokens. Similar observation applies to informative tokens 10 and 12, which clearly reflects that our method is capable of capturing information of critical locations such as eyes, noses and key parts of body and aggregating information to the most.  

It is worth emphasizing that informative regions are not necessarily foreground areas. For example, the sky and mountains in the distance also receive much attention in the image. Instead of paying excessive focus on foreground areas, we treat them as the same and reserve them adaptively, which is of great significance to dense prediction tasks. 

\section{Conclusion}
In this paper, we rethink token reduction as matrix transformation and unify existing token compression methods as special forms of matrix construction. Considering token-exclusive property of existing methods that leads to sub-optimal performance, we propose a general Token Transforming framework that allows more flexible many-to-many transformation and introduce an algorithm which can reserve token information to the most in a training-free manner.
Besides image classification tasks, we extend our method to numerous dense prediction tasks including semantic segmentation, object detection, depth estimation and large model generation to validate the superiority, efficiency, scalability and generalizability of our method.

{
    \small
    \bibliographystyle{ieeenat_fullname}
    \bibliography{main}

\begin{thebibliography}{56}
\providecommand{\natexlab}[1]{#1}
\providecommand{\url}[1]{\texttt{#1}}
\expandafter\ifx\csname urlstyle\endcsname\relax
  \providecommand{\doi}[1]{doi: #1}\else
  \providecommand{\doi}{doi: \begingroup \urlstyle{rm}\Url}\fi

\bibitem[Bolya et~al.(2022)Bolya, Fu, Dai, Zhang, Feichtenhofer, and Hoffman]{bolya2022token}
Daniel Bolya, Cheng-Yang Fu, Xiaoliang Dai, Peizhao Zhang, Christoph Feichtenhofer, and Judy Hoffman.
\newblock Token merging: Your vit but faster.
\newblock \emph{arXiv preprint arXiv:2210.09461}, 2022.

\bibitem[Chen et~al.(2021)Chen, Fan, and Panda]{chen2021crossvit}
Chun-Fu~Richard Chen, Quanfu Fan, and Rameswar Panda.
\newblock Crossvit: Cross-attention multi-scale vision transformer for image classification.
\newblock In \emph{Proceedings of the IEEE/CVF international conference on computer vision}, pages 357--366, 2021.

\bibitem[Cordts et~al.(2016)Cordts, Omran, Ramos, Rehfeld, Enzweiler, Benenson, Franke, Roth, and Schiele]{cordts2016cityscapes}
Marius Cordts, Mohamed Omran, Sebastian Ramos, Timo Rehfeld, Markus Enzweiler, Rodrigo Benenson, Uwe Franke, Stefan Roth, and Bernt Schiele.
\newblock The cityscapes dataset for semantic urban scene understanding.
\newblock In \emph{Proceedings of the IEEE conference on computer vision and pattern recognition}, pages 3213--3223, 2016.

\bibitem[Deng et~al.(2009)Deng, Dong, Socher, Li, Li, and Fei-Fei]{deng2009imagenet}
Jia Deng, Wei Dong, Richard Socher, Li-Jia Li, Kai Li, and Li Fei-Fei.
\newblock Imagenet: A large-scale hierarchical image database.
\newblock In \emph{2009 IEEE conference on computer vision and pattern recognition}, pages 248--255. Ieee, 2009.

\bibitem[Dosovitskiy et~al.(2020)Dosovitskiy, Beyer, Kolesnikov, Weissenborn, Zhai, Unterthiner, Dehghani, Minderer, Heigold, Gelly, et~al.]{dosovitskiy2020image}
Alexey Dosovitskiy, Lucas Beyer, Alexander Kolesnikov, Dirk Weissenborn, Xiaohua Zhai, Thomas Unterthiner, Mostafa Dehghani, Matthias Minderer, Georg Heigold, Sylvain Gelly, et~al.
\newblock An image is worth 16x16 words: Transformers for image recognition at scale.
\newblock \emph{arXiv preprint arXiv:2010.11929}, 2020.

\bibitem[Fayyaz et~al.(2022)Fayyaz, Koohpayegani, Jafari, Sengupta, Joze, Sommerlade, Pirsiavash, and Gall]{fayyaz2022adaptive}
Mohsen Fayyaz, Soroush~Abbasi Koohpayegani, Farnoush~Rezaei Jafari, Sunando Sengupta, Hamid Reza~Vaezi Joze, Eric Sommerlade, Hamed Pirsiavash, and J{\"u}rgen Gall.
\newblock Adaptive token sampling for efficient vision transformers.
\newblock In \emph{Computer Vision--ECCV 2022: 17th European Conference, Tel Aviv, Israel, October 23--27, 2022, Proceedings, Part XI}, pages 396--414. Springer, 2022.

\bibitem[Gao et~al.(2018)Gao, Zhao, Dudziak, Mullins, and Xu]{gao2018dynamic}
Xitong Gao, Yiren Zhao, {\L}ukasz Dudziak, Robert Mullins, and Cheng-zhong Xu.
\newblock Dynamic channel pruning: Feature boosting and suppression.
\newblock \emph{arXiv preprint arXiv:1810.05331}, 2018.

\bibitem[Goyal et~al.(2017)Goyal, Khot, Summers-Stay, Batra, and Parikh]{goyal2017making}
Yash Goyal, Tejas Khot, Douglas Summers-Stay, Dhruv Batra, and Devi Parikh.
\newblock Making the v in vqa matter: Elevating the role of image understanding in visual question answering.
\newblock In \emph{Proceedings of the IEEE conference on computer vision and pattern recognition}, pages 6904--6913, 2017.

\bibitem[Han et~al.(2021)Han, Huang, Song, Yang, Wang, and Wang]{han2021dynamic}
Yizeng Han, Gao Huang, Shiji Song, Le Yang, Honghui Wang, and Yulin Wang.
\newblock Dynamic neural networks: A survey.
\newblock \emph{IEEE Transactions on Pattern Analysis and Machine Intelligence}, 44\penalty0 (11):\penalty0 7436--7456, 2021.

\bibitem[He et~al.(2022)He, Chen, Xie, Li, Doll{\'a}r, and Girshick]{he2022masked}
Kaiming He, Xinlei Chen, Saining Xie, Yanghao Li, Piotr Doll{\'a}r, and Ross Girshick.
\newblock Masked autoencoders are scalable vision learners.
\newblock In \emph{Proceedings of the IEEE/CVF conference on computer vision and pattern recognition}, pages 16000--16009, 2022.

\bibitem[Hinton et~al.(2015)Hinton, Vinyals, and Dean]{hinton2015distilling}
Geoffrey Hinton, Oriol Vinyals, and Jeff Dean.
\newblock Distilling the knowledge in a neural network.
\newblock \emph{arXiv preprint arXiv:1503.02531}, 2015.

\bibitem[Jiang et~al.(2021)Jiang, Hou, Yuan, Zhou, Shi, Jin, Wang, and Feng]{jiang2021all}
Zi-Hang Jiang, Qibin Hou, Li Yuan, Daquan Zhou, Yujun Shi, Xiaojie Jin, Anran Wang, and Jiashi Feng.
\newblock All tokens matter: Token labeling for training better vision transformers.
\newblock \emph{Advances in neural information processing systems}, 34:\penalty0 18590--18602, 2021.

\bibitem[Kong et~al.(2022)Kong, Dong, Ma, Meng, Niu, Sun, Shen, Yuan, Ren, Tang, et~al.]{kong2022spvit}
Zhenglun Kong, Peiyan Dong, Xiaolong Ma, Xin Meng, Wei Niu, Mengshu Sun, Xuan Shen, Geng Yuan, Bin Ren, Hao Tang, et~al.
\newblock Spvit: Enabling faster vision transformers via latency-aware soft token pruning.
\newblock In \emph{European conference on computer vision}, pages 620--640. Springer, 2022.

\bibitem[Kong et~al.(2023)Kong, Ma, Yuan, Sun, Xie, Dong, Meng, Shen, Tang, Qin, et~al.]{kong2023peeling}
Zhenglun Kong, Haoyu Ma, Geng Yuan, Mengshu Sun, Yanyue Xie, Peiyan Dong, Xin Meng, Xuan Shen, Hao Tang, Minghai Qin, et~al.
\newblock Peeling the onion: Hierarchical reduction of data redundancy for efficient vision transformer training.
\newblock In \emph{Proceedings of the AAAI Conference on Artificial Intelligence}, pages 8360--8368, 2023.

\bibitem[Kong et~al.(2025)Kong, Li, Zeng, Xin, Messica, Lin, Zhao, Kellis, Tang, and Zitnik]{kong2025token}
Zhenglun Kong, Yize Li, Fanhu Zeng, Lei Xin, Shvat Messica, Xue Lin, Pu Zhao, Manolis Kellis, Hao Tang, and Marinka Zitnik.
\newblock Token reduction should go beyond efficiency in generative models--from vision, language to multimodality.
\newblock \emph{arXiv preprint arXiv:2505.18227}, 2025.

\bibitem[Lee et~al.(2022)Lee, Kim, Willette, and Hwang]{lee2022mpvit}
Youngwan Lee, Jonghee Kim, Jeffrey Willette, and Sung~Ju Hwang.
\newblock Mpvit: Multi-path vision transformer for dense prediction.
\newblock In \emph{Proceedings of the IEEE/CVF Conference on Computer Vision and Pattern Recognition}, pages 7287--7296, 2022.

\bibitem[Li et~al.(2024)Li, Ma, Yang, and Yang]{li2024vidtome}
Xirui Li, Chao Ma, Xiaokang Yang, and Ming-Hsuan Yang.
\newblock Vidtome: Video token merging for zero-shot video editing.
\newblock In \emph{Proceedings of the IEEE/CVF Conference on Computer Vision and Pattern Recognition}, 2024.

\bibitem[Li et~al.(2022)Li, Mao, Girshick, and He]{li2022exploring}
Yanghao Li, Hanzi Mao, Ross Girshick, and Kaiming He.
\newblock Exploring plain vision transformer backbones for object detection.
\newblock In \emph{European Conference on Computer Vision}, pages 280--296. Springer, 2022.

\bibitem[Li et~al.(2023)Li, Du, Zhou, Wang, Zhao, and Wen]{li2023evaluating}
Yifan Li, Yifan Du, Kun Zhou, Jinpeng Wang, Wayne~Xin Zhao, and Ji-Rong Wen.
\newblock Evaluating object hallucination in large vision-language models.
\newblock In \emph{Proceedings of the 2023 Conference on Empirical Methods in Natural Language Processing}, pages 292--305, 2023.

\bibitem[Liang et~al.(2022{\natexlab{a}})Liang, Yuan, Ding, Luo, Lin, Jia, Zhang, Zhang, and Hu]{liang2022expediting}
Weicong Liang, Yuhui Yuan, Henghui Ding, Xiao Luo, Weihong Lin, Ding Jia, Zheng Zhang, Chao Zhang, and Han Hu.
\newblock Expediting large-scale vision transformer for dense prediction without fine-tuning.
\newblock \emph{Advances in Neural Information Processing Systems}, 35:\penalty0 35462--35477, 2022{\natexlab{a}}.

\bibitem[Liang et~al.(2022{\natexlab{b}})Liang, Ge, Tong, Song, Wang, and Xie]{liang2022not}
Youwei Liang, Chongjian Ge, Zhan Tong, Yibing Song, Jue Wang, and Pengtao Xie.
\newblock Not all patches are what you need: Expediting vision transformers via token reorganizations.
\newblock \emph{arXiv preprint arXiv:2202.07800}, 2022{\natexlab{b}}.

\bibitem[Lin et~al.(2014)Lin, Maire, Belongie, Hays, Perona, Ramanan, Doll{\'a}r, and Zitnick]{lin2014coco}
Tsung-Yi Lin, Michael Maire, Serge Belongie, James Hays, Pietro Perona, Deva Ramanan, Piotr Doll{\'a}r, and C~Lawrence Zitnick.
\newblock Microsoft coco: Common objects in context.
\newblock In \emph{Computer Vision--ECCV 2014: 13th European Conference, Zurich, Switzerland, September 6-12, 2014, Proceedings, Part V 13}, pages 740--755. Springer, 2014.

\bibitem[Liu et~al.(2024)Liu, Li, Wu, and Lee]{liu2024visual}
Haotian Liu, Chunyuan Li, Qingyang Wu, and Yong~Jae Lee.
\newblock Visual instruction tuning.
\newblock \emph{Advances in neural information processing systems}, 36, 2024.

\bibitem[Liu et~al.(2022)Liu, Wu, and Guo]{liu2022adaptive}
Xiangcheng Liu, Tianyi Wu, and Guodong Guo.
\newblock Adaptive sparse vit: Towards learnable adaptive token pruning by fully exploiting self-attention.
\newblock \emph{arXiv preprint arXiv:2209.13802}, 2022.

\bibitem[Liu et~al.(2021)Liu, Lin, Cao, Hu, Wei, Zhang, Lin, and Guo]{liu2021swin}
Ze Liu, Yutong Lin, Yue Cao, Han Hu, Yixuan Wei, Zheng Zhang, Stephen Lin, and Baining Guo.
\newblock Swin transformer: Hierarchical vision transformer using shifted windows.
\newblock In \emph{Proceedings of the IEEE/CVF international conference on computer vision}, pages 10012--10022, 2021.

\bibitem[Long et~al.(2023)Long, Zhao, Pi, Wang, and Wang]{long2023beyond}
Sifan Long, Zhen Zhao, Jimin Pi, Shengsheng Wang, and Jingdong Wang.
\newblock Beyond attentive tokens: Incorporating token importance and diversity for efficient vision transformers.
\newblock In \emph{Proceedings of the IEEE/CVF Conference on Computer Vision and Pattern Recognition}, pages 10334--10343, 2023.

\bibitem[Lu et~al.(2022)Lu, Mishra, Xia, Qiu, Chang, Zhu, Tafjord, Clark, and Kalyan]{lu2022learn}
Pan Lu, Swaroop Mishra, Tanglin Xia, Liang Qiu, Kai-Wei Chang, Song-Chun Zhu, Oyvind Tafjord, Peter Clark, and Ashwin Kalyan.
\newblock Learn to explain: Multimodal reasoning via thought chains for science question answering.
\newblock \emph{Advances in Neural Information Processing Systems}, 35:\penalty0 2507--2521, 2022.

\bibitem[Molchanov et~al.(2019)Molchanov, Mallya, Tyree, Frosio, and Kautz]{molchanov2019importance}
Pavlo Molchanov, Arun Mallya, Stephen Tyree, Iuri Frosio, and Jan Kautz.
\newblock Importance estimation for neural network pruning.
\newblock In \emph{Proceedings of the IEEE/CVF conference on computer vision and pattern recognition}, pages 11264--11272, 2019.

\bibitem[Radford et~al.(2018)Radford, Narasimhan, Salimans, Sutskever, et~al.]{radford2018improving}
Alec Radford, Karthik Narasimhan, Tim Salimans, Ilya Sutskever, et~al.
\newblock Improving language understanding by generative pre-training.
\newblock 2018.

\bibitem[Ranftl et~al.(2021)Ranftl, Bochkovskiy, and Koltun]{ranftl2021vision}
Ren{\'e} Ranftl, Alexey Bochkovskiy, and Vladlen Koltun.
\newblock Vision transformers for dense prediction.
\newblock In \emph{Proceedings of the IEEE/CVF international conference on computer vision}, pages 12179--12188, 2021.

\bibitem[Rao et~al.(2021)Rao, Zhao, Liu, Lu, Zhou, and Hsieh]{rao2021dynamicvit}
Yongming Rao, Wenliang Zhao, Benlin Liu, Jiwen Lu, Jie Zhou, and Cho-Jui Hsieh.
\newblock Dynamicvit: Efficient vision transformers with dynamic token sparsification.
\newblock \emph{Advances in neural information processing systems}, 34:\penalty0 13937--13949, 2021.

\bibitem[Ren et~al.(2022)Ren, Li, Wang, Xiao, Du, Liang, and Chang]{ren2022beyond}
Pengzhen Ren, Changlin Li, Guangrun Wang, Yun Xiao, Qing Du, Xiaodan Liang, and Xiaojun Chang.
\newblock Beyond fixation: Dynamic window visual transformer.
\newblock In \emph{Proceedings of the IEEE/CVF Conference on Computer Vision and Pattern Recognition}, pages 11987--11997, 2022.

\bibitem[Ryoo et~al.(2021)Ryoo, Piergiovanni, Arnab, Dehghani, and Angelova]{ryoo2021tokenlearner}
Michael Ryoo, AJ Piergiovanni, Anurag Arnab, Mostafa Dehghani, and Anelia Angelova.
\newblock Tokenlearner: Adaptive space-time tokenization for videos.
\newblock \emph{Advances in Neural Information Processing Systems}, 34:\penalty0 12786--12797, 2021.

\bibitem[Shang et~al.(2024)Shang, Cai, Xu, Lee, and Yan]{shang2024prumerge}
Yuzhang Shang, Mu Cai, Bingxin Xu, Yong~Jae Lee, and Yan Yan.
\newblock Llava-prumerge: Adaptive token reduction for efficient large multimodal models.
\newblock \emph{arXiv preprint arXiv:2403.15388}, 2024.

\bibitem[Silberman et~al.(2012)Silberman, Hoiem, Kohli, and Fergus]{silberman2012indoor}
Nathan Silberman, Derek Hoiem, Pushmeet Kohli, and Rob Fergus.
\newblock Indoor segmentation and support inference from rgbd images.
\newblock In \emph{Computer Vision--ECCV 2012: 12th European Conference on Computer Vision, Florence, Italy, October 7-13, 2012, Proceedings, Part V 12}, pages 746--760. Springer, 2012.

\bibitem[Singh et~al.(2019)Singh, Natarajan, Shah, Jiang, Chen, Batra, Parikh, and Rohrbach]{singh2019towards}
Amanpreet Singh, Vivek Natarajan, Meet Shah, Yu Jiang, Xinlei Chen, Dhruv Batra, Devi Parikh, and Marcus Rohrbach.
\newblock Towards vqa models that can read.
\newblock In \emph{Proceedings of the IEEE/CVF conference on computer vision and pattern recognition}, pages 8317--8326, 2019.

\bibitem[Steiner et~al.(2022)Steiner, Kolesnikov, Zhai, Wightman, Uszkoreit, and Beyer]{steiner2022train}
Andreas~Peter Steiner, Alexander Kolesnikov, Xiaohua Zhai, Ross Wightman, Jakob Uszkoreit, and Lucas Beyer.
\newblock How to train your vit? data, augmentation, and regularization in vision transformers.
\newblock 2022.

\bibitem[Strudel et~al.(2021)Strudel, Garcia, Laptev, and Schmid]{strudel2021segmenter}
Robin Strudel, Ricardo Garcia, Ivan Laptev, and Cordelia Schmid.
\newblock Segmenter: Transformer for semantic segmentation.
\newblock In \emph{Proceedings of the IEEE/CVF international conference on computer vision}, pages 7262--7272, 2021.

\bibitem[Sui et~al.(2021)Sui, Yin, Xie, Phan, Aliari~Zonouz, and Yuan]{sui2021chip}
Yang Sui, Miao Yin, Yi Xie, Huy Phan, Saman Aliari~Zonouz, and Bo Yuan.
\newblock Chip: Channel independence-based pruning for compact neural networks.
\newblock \emph{Advances in Neural Information Processing Systems}, 34:\penalty0 24604--24616, 2021.

\bibitem[Sun et~al.(2017)Sun, Shrivastava, Singh, and Gupta]{sun2017revisiting}
Chen Sun, Abhinav Shrivastava, Saurabh Singh, and Abhinav Gupta.
\newblock Revisiting unreasonable effectiveness of data in deep learning era.
\newblock In \emph{Proceedings of the IEEE international conference on computer vision}, pages 843--852, 2017.

\bibitem[Tang et~al.(2022)Tang, Han, Wang, Xu, Guo, Xu, and Tao]{tang2022patch}
Yehui Tang, Kai Han, Yunhe Wang, Chang Xu, Jianyuan Guo, Chao Xu, and Dacheng Tao.
\newblock Patch slimming for efficient vision transformers.
\newblock In \emph{Proceedings of the IEEE/CVF Conference on Computer Vision and Pattern Recognition}, pages 12165--12174, 2022.

\bibitem[Touvron et~al.(2021)Touvron, Cord, Douze, Massa, Sablayrolles, and J{\'e}gou]{touvron2021training}
Hugo Touvron, Matthieu Cord, Matthijs Douze, Francisco Massa, Alexandre Sablayrolles, and Herv{\'e} J{\'e}gou.
\newblock Training data-efficient image transformers \& distillation through attention.
\newblock In \emph{International conference on machine learning}, pages 10347--10357. PMLR, 2021.

\bibitem[Vaswani et~al.(2017)Vaswani, Shazeer, Parmar, Uszkoreit, Jones, Gomez, Kaiser, and Polosukhin]{vaswani2017attention}
Ashish Vaswani, Noam Shazeer, Niki Parmar, Jakob Uszkoreit, Llion Jones, Aidan~N Gomez, {\L}ukasz Kaiser, and Illia Polosukhin.
\newblock Attention is all you need.
\newblock \emph{Advances in neural information processing systems}, 30, 2017.

\bibitem[Wang et~al.(2021)Wang, Huang, Song, Huang, and Huang]{wang2021not}
Yulin Wang, Rui Huang, Shiji Song, Zeyi Huang, and Gao Huang.
\newblock Not all images are worth 16x16 words: Dynamic transformers for efficient image recognition.
\newblock \emph{Advances in Neural Information Processing Systems}, 34:\penalty0 11960--11973, 2021.

\bibitem[Wang et~al.(2022)Wang, Liu, Huang, Chen, Zhang, Lin, and Wang]{wang2022qsfm}
Zidu Wang, Xuexin Liu, Long Huang, Yunqing Chen, Yufei Zhang, Zhikang Lin, and Rui Wang.
\newblock Qsfm: Model pruning based on quantified similarity between feature maps for ai on edge.
\newblock \emph{IEEE Internet of Things Journal}, 9\penalty0 (23):\penalty0 24506--24515, 2022.

\bibitem[Wei et~al.(2023)Wei, Ye, Zhang, Tang, and Liang]{wei2023joint}
Siyuan Wei, Tianzhu Ye, Shen Zhang, Yao Tang, and Jiajun Liang.
\newblock Joint token pruning and squeezing towards more aggressive compression of vision transformers.
\newblock In \emph{Proceedings of the IEEE/CVF Conference on Computer Vision and Pattern Recognition}, pages 2092--2101, 2023.

\bibitem[Wu et~al.(2023)Wu, Zeng, Wang, and Chen]{wu2023ppt}
Xinjian Wu, Fanhu Zeng, Xiudong Wang, and Xinghao Chen.
\newblock Ppt: Token pruning and pooling for efficient vision transformers.
\newblock \emph{arXiv preprint arXiv:2310.01812}, 2023.

\bibitem[Xu et~al.(2022)Xu, Zhang, Zhang, Sheng, Li, Dong, Zhang, Xu, and Sun]{xu2022evo}
Yifan Xu, Zhijie Zhang, Mengdan Zhang, Kekai Sheng, Ke Li, Weiming Dong, Liqing Zhang, Changsheng Xu, and Xing Sun.
\newblock Evo-vit: Slow-fast token evolution for dynamic vision transformer.
\newblock In \emph{Proceedings of the AAAI Conference on Artificial Intelligence}, pages 2964--2972, 2022.

\bibitem[Yang et~al.(2023)Yang, Yin, Shen, Molchanov, Li, and Kautz]{yang2023global}
Huanrui Yang, Hongxu Yin, Maying Shen, Pavlo Molchanov, Hai Li, and Jan Kautz.
\newblock Global vision transformer pruning with hessian-aware saliency.
\newblock In \emph{Proceedings of the IEEE/CVF Conference on Computer Vision and Pattern Recognition}, pages 18547--18557, 2023.

\bibitem[Yin et~al.(2022)Yin, Vahdat, Alvarez, Mallya, Kautz, and Molchanov]{yin2022vit}
Hongxu Yin, Arash Vahdat, Jose~M Alvarez, Arun Mallya, Jan Kautz, and Pavlo Molchanov.
\newblock A-vit: Adaptive tokens for efficient vision transformer.
\newblock In \emph{Proceedings of the IEEE/CVF Conference on Computer Vision and Pattern Recognition}, pages 10809--10818, 2022.

\bibitem[Yuan et~al.(2021)Yuan, Chen, Wang, Yu, Shi, Jiang, Tay, Feng, and Yan]{yuan2021tokens}
Li Yuan, Yunpeng Chen, Tao Wang, Weihao Yu, Yujun Shi, Zi-Hang Jiang, Francis~EH Tay, Jiashi Feng, and Shuicheng Yan.
\newblock Tokens-to-token vit: Training vision transformers from scratch on imagenet.
\newblock In \emph{Proceedings of the IEEE/CVF international conference on computer vision}, pages 558--567, 2021.

\bibitem[Yue et~al.(2021)Yue, Sun, Kuang, Wei, Torr, Zhang, and Lin]{yue2021vision}
Xiaoyu Yue, Shuyang Sun, Zhanghui Kuang, Meng Wei, Philip~HS Torr, Wayne Zhang, and Dahua Lin.
\newblock Vision transformer with progressive sampling.
\newblock In \emph{Proceedings of the IEEE/CVF International Conference on Computer Vision}, pages 387--396, 2021.

\bibitem[Zeng and Yu()]{zeng2024m2m}
Fanhu Zeng and Deli Yu.
\newblock M2m-tag: Training-free many-to-many token aggregation for vision transformer acceleration.
\newblock In \emph{Workshop on Machine Learning and Compression, NeurIPS 2024}.

\bibitem[Zeng et~al.(2025)Zeng, Guo, Zhu, Shen, and Tang]{zeng2025parameter}
Fanhu Zeng, Haiyang Guo, Fei Zhu, Li Shen, and Hao Tang.
\newblock Parameter efficient merging for multimodal large language models with complementary parameter adaptation.
\newblock \emph{arXiv preprint arXiv:2502.17159}, 2025.

\bibitem[Zhang et~al.(2022)Zhang, Li, Liu, Zhang, Su, Zhu, Ni, and Shum]{zhang2022dino}
Hao Zhang, Feng Li, Shilong Liu, Lei Zhang, Hang Su, Jun Zhu, Lionel~M Ni, and Heung-Yeung Shum.
\newblock Dino: Detr with improved denoising anchor boxes for end-to-end object detection.
\newblock \emph{arXiv preprint arXiv:2203.03605}, 2022.

\bibitem[Zhou et~al.(2019)Zhou, Zhao, Puig, Xiao, Fidler, Barriuso, and Torralba]{zhou2019semantic}
Bolei Zhou, Hang Zhao, Xavier Puig, Tete Xiao, Sanja Fidler, Adela Barriuso, and Antonio Torralba.
\newblock Semantic understanding of scenes through the ade20k dataset.
\newblock \emph{International Journal of Computer Vision}, 127:\penalty0 302--321, 2019.

\end{thebibliography}
}

\end{document}